\documentclass[a4paper]{article}

\usepackage{INTERSPEECH2021}
\newcommand{\myparagraph}[1]{\vspace{.05in}\noindent{\bf{#1}}~~}

\title{Contextualized Attention-based Knowledge Transfer for Spoken Conversational Question Answering}

\name{Chenyu You$^{1 \dagger}$\thanks{$^{\dagger}$ Indicates equal contribution},  Nuo Chen$^{2 \dagger}$, Yuexian Zou$^{2,3,*}$\thanks{$^{*}$ Corresponding Author}}
\address{$^{1}$Department of Electrical Engineering, Yale University, CT, USA\\
$^{2}$ADSPLAB, School of ECE, Peking University, Shenzhen, China\\
$^3$Peng Cheng Laboratory, Shenzhen, China
}

\email{chenyu.you@yale.edu, \{nuochen,zouyx\}@pku.edu.cn}

\begin{document}

\maketitle
\begin{abstract}
    Spoken conversational question answering (SCQA) requires machines to model the flow of multi-turn conversation given the speech utterances and text corpora. Different from traditional text question answering (QA) tasks, SCQA involves audio signal processing, passage comprehension, and contextual understanding. However, ASR systems introduce unexpected noisy signals to the transcriptions, which result in performance degradation on SCQA. To overcome the problem, we propose CADNet, a novel contextualized attention-based distillation approach, which applies both cross-attention and self-attention to obtain ASR-robust contextualized embedding representations of the passage and dialogue history for performance improvements. We also introduce the spoken conventional knowledge distillation framework to distill the ASR-robust knowledge from the estimated probabilities of the~\textit{teacher} model to the~\textit{student}. We conduct extensive experiments on the Spoken-CoQA dataset and demonstrate that our approach achieves remarkable performance in this task.

\end{abstract}
\noindent\textbf{Index Terms}: spoken conversational question answering, machine reading comprehension, conversational question answering

\section{Introduction}
\label{sec:intro}
Neural network based end-to-end methods \cite{you2021momentum,you2019ct,you2020unsupervised,you2018structurally,you2019low} have attracted a lot of attention in the machine learning community. With the recent advances in machine learning, spoken question answering (SQA) has become an important research topic during the past few years. To be specific, SQA requires the machine to fully understand the spoken content of the document and questions, and then predict an answer. A major limitation for this field is the lack of benchmark datasets. To alleviate such issue, several benchmark datasets~\cite{lee2015spoken,li2018spoken,lee2018odsqa,you2020data} are released to the speech processing and natural language processing communities. Spoken-SQuAD~\cite{li2018spoken} is one of the typical benchmarks, which uses CMU Sphinx to generate auto-transcribed text given the Text-SQuAD \cite{rajpurkar2016squad} dataset. However, the SQA tasks only share the single-turn setting to answer a single question given the spoken document, which is far from real SQA scenarios. For example, in a real-world interview, the discourse structure is more complex, which including multi-part conversation. Thus, it is crucial to enable the QA systems to address the turn structure (e.g., meetings, debate) and the consistent discourse interpretation.

\begin{table}[t]
\caption{An example from Spoken-CoQA. We can observe large misalignment between the manual transcripts and the corresponding ASR transcripts. Note that the misalignment is in~\textbf{bold} font.}
\label{table1}
\begin{center}
\begin{tabular}{ll}
\multicolumn{1}{c}{\bf Manual Transcript}  &\multicolumn{1}{c}{\bf ASR Transcript}
\\ \hline
\begin{minipage}[t]{0.45\columnwidth}%
Once there was a beautiful fish named Asta. Asta lived in the ocean. There were lots of other fish in the ocean where Asta lived. They played all day long. One day, a bottle floated by over the heads of Asta and his friends. They looked up and saw the bottle \dots
\end{minipage}
&
\begin{minipage}[t]{0.45\columnwidth}%
once there was a beautiful fish named \textbf{After}. \textbf{After} lived in the ocean. There were lots of other fish in the ocean \textbf{we're} asked to live. They played all day long. One day, a bottle floated by over the heads of \textbf{vast} and his friends. They looked up and saw the bottle \dots
\end{minipage}
\\
\\
\begin{minipage}[t]{0.45\columnwidth}%
Q$_{1}$: What was the name of the fish? \\
A$_{1}$: Asta  \\
R$_{1}$:Asta.
\end{minipage}
&
\begin{minipage}[t]{0.45\columnwidth}%
ASR-Q$_{1}$: What was the name of the fish? \\
A$_{1}$: \textbf{After}  \\
R$_{1}$: \textbf{After}.
\end{minipage}
\\ \\
\begin{minipage}[t]{0.45\columnwidth}%
Q$_{2}$: What looked like a birds belly? \\
A$_{2}$: a bottle  \\
R$_{2}$: a bottle
\end{minipage}
&
\begin{minipage}[t]{0.45\columnwidth}%
ASR-Q$_{2}$: What looked like a bird to \textbf{delhi}?    \\
A$_{2}$: a bottle  \\
R$_{2}$: a bottle 
\end{minipage}
\\ \bottomrule
\end{tabular}
\end{center}
\end{table}

SQA includes audio signal processing, passage comprehension, and contextual understanding. The typical way is to translate speech content into text forms, and then apply the state-of-the-art Text-QA models on the transcribed text documents~\cite{lee2018odsqa,luo2019spoken,lee2019mitigating,su2020improving,you2020knowledge,tsai2019MULT,chen2020adaptive,SunSSL20,unlu2021uncertainty,peskov2019mitigating,siriwardhana2020jointly,chen2021selfsupervised,peskov2019mitigating,arisoy2021uncertainty}. However, the recent study~\cite{tseng2016towards} suggests that ASR errors may severely affect answering accuracy. Previous works have been proposed to address such issues. Lee et al.~\cite{lee2018odsqa} shows that using sub-word units yields promising improvements in terms of accuracy. Very recently, domain adversarial learning~\cite{lee2019mitigating} was introduced to mitigate the effects of ASR errors effectively. Most recently, You et al.~\cite{you2020data} released the first SCQA benchmark dataset - Spoken-CoQA, and then propose to use a~\textit{teacher-student} paradigm to boost the network performance on highly noisy ASR transcripts.

\begin{figure*}[t]
    \centering
    \includegraphics[width=\linewidth]{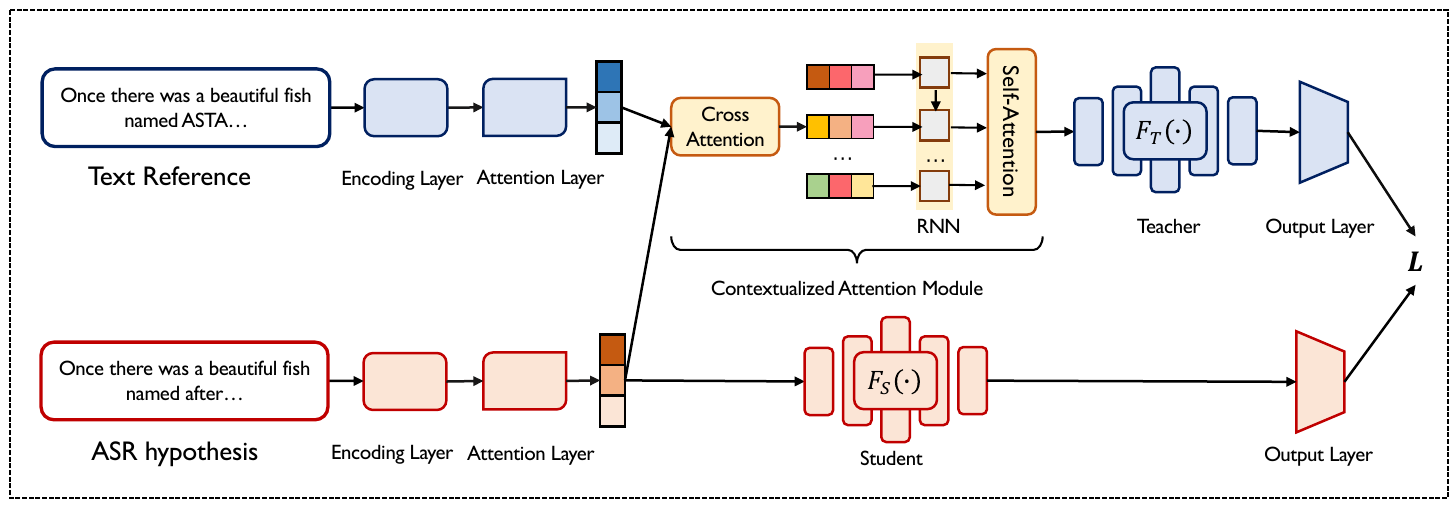}
    \caption{Overview of our proposed method.}
    \label{fig:i}
\end{figure*}

In this paper, based on our previous work~\cite{you2020data} on the spoken conversational question answering task, we present CADNet, a contextual attention-based data distillation network. Our network first leverages both cross-attention and self-attention to extract relevant information between auto-transcribed (ASR) texts and the reference text documents to better understand the corpus of spoken documents and questions effectively. Then we distill reliable supervision signals from the reference written documents, and uses these predictions to guide the training of the~\textit{student}. We evaluate our method on the Spoken-CoQA dataset, and experimental results demonstrate that our method exhibits good improvements in terms of accuracy over several state-of-the-art models.

\section{Dataset}
% \subsection{Dataset}
We use the listening comprehension benchmark dataset Spoken-CoQA \cite{you2020data} corpus. Each example in this dataset is defined as follows: \{$P_{i},Q_{i},A_{i}\}_{1}^{N}$, where  $P_{i}$ denotes the given passage. $Q_{i}$=\{$q_{i1},q_{i2},...,q_{iL}$\} and $ A_{i}$= \{$a_{i1},a_{i2},...,a_{iL}$\}  represent a passage with $L$-turn queries and corresponding answers, respectively. Given a passage $P_{i}$ and multi-turn history questions \{${q_{i1},q_{i2},...,q_{iL-1}}$\} and answers \{${a_{i1},a_{i2},...,}$ $a_{iL-1}$\}, our goal is to generate $a_{iL}$ for the given current question $q_{iL}$. Note that questions and documents in Spoken-CoQA are in both text and spoken forms, and answers are in the text form. Table \ref{table1} is an example selected from the Spoken-CoQA development set. As we can see, given the text document (ASR-document) the conversation begins with the question $Q_1$ (ASR-$Q_1$), then the Text-QA model needs to answer $Q_1$ (ASR-$Q_1$) with $A_1$ based on a contiguous text span $R_1$. This suggests that ASR transcripts (both the document and questions) are more difficult for the model to comprehend, reason and even predict correct answers. The word error rate (WER) is 18.7$\%$.

\section{Method}
\subsection{Model Overview}

In this work, we focus primarily on applying the existing Text-based Conversational Machine Reading Comprehension (Text-CMRC) models to handle highly noisy transcriptions from the ASR systems. Generally, the Text-CMRC models include three major parts: encoding layer, attention layer, and output layer.~\textit{Encoding Layers} uses the documents and conversations (questions and answers) to encode the word into the corresponding feature embedding (e.g., character embedding, word embedding, and contextual embedding). In the \textit{Attention Layer}, we then extract the most relevant information from the context for answering the question by condensing the context representations of documents into a fixed-length vector. Finally, \textit{Output Layer} predicts an answer for the current question given the learned representations.

To mitigate the adverse effects caused by the ASR errors in the SCQA task, we propose a novel contextualized attention-based distillation network (CADNet). We first introduce the cross-attention mechanism to align the mismatch between the reference manual transcript and the corresponding ASR transcripts. We then use multi-layer recurrent neural networks (RNN) to learn ASR-robust contextual representations from the document collections. Next, we conduct self-attention to establish the correlations between words at different positions and capture additional cues from the manual transcripts. Finally, we improve the knowledge transfer by training a robust ~\textit{teacher} model and then enroll the improved knowledge into \textit{student} model trained on ASR transcriptions to provide better performance improvements. We present the proposed method in Figure\ref{fig:i}.

\begin{table*}[ht]
    \caption{Comparison of four baselines. Note that we denote Text-CoQA and Spoken-CoQA test set as T-CoQA and S-CoQA test for brevity.}
    % \footnotesize
    \centering
    \begin{tabular}{l | c c|c c|c c|c c}

    % \bottomrule[1pt]
    % \hline\hline
        &\multicolumn{4}{c}{\textbf{T-CoQA}}&
        \multicolumn{4}{c}{\textbf{S-CoQA}}\\
        \hline\hline
        &\multicolumn{2}{c}{\textbf{T-CoQA dev}}&
        \multicolumn{2}{|c}{\textbf{S-CoQA test}}&\multicolumn{2}{|c|}{\textbf{T-CoQA dev}}&
        \multicolumn{2}{|c}{\textbf{S-CoQA test}}\\
        % \hline
        \textbf{Methods} &EM &F1 &EM &F1 & EM&F1 &EM&F1 \\
        \hline
        FlowQA \cite{huang2018flowqa} & 66.8& 75.1 & 44.1& 56.8 & 40.9& 51.6 & 22.1& 34.7  \\
        SDNet \cite{zhu2018sdnet} & 68.1 & 76.9  & 39.5 & 51.2 & 40.1 & 52.5  & 41.5 & 53.1\\
        BERT-base~\cite{devlin2018bert} & 67.7 & 77.7  & 41.8 & 54.7 & 42.3 & 55.8  & 40.6 & 54.1\\
        ALBERT-base~\cite{lan2019albert}  & 71.4&80.6 &42.6& 54.8 & 42.7&56.0 &41.4& 55.2\\
        \hline
        Average  & 68.5 &77.6 &42& 54.4 & 41.5 & 54.0 & 36.4 & 49.3\\
        % \bottomrule
    \end{tabular}
    \label{tab:my_label_2}
    % \vspace{-10pt}
\end{table*}

\subsection{Contextualized Attention (CA) Module}
\begin{figure}
    \centering
    \includegraphics[width=0.8\linewidth]{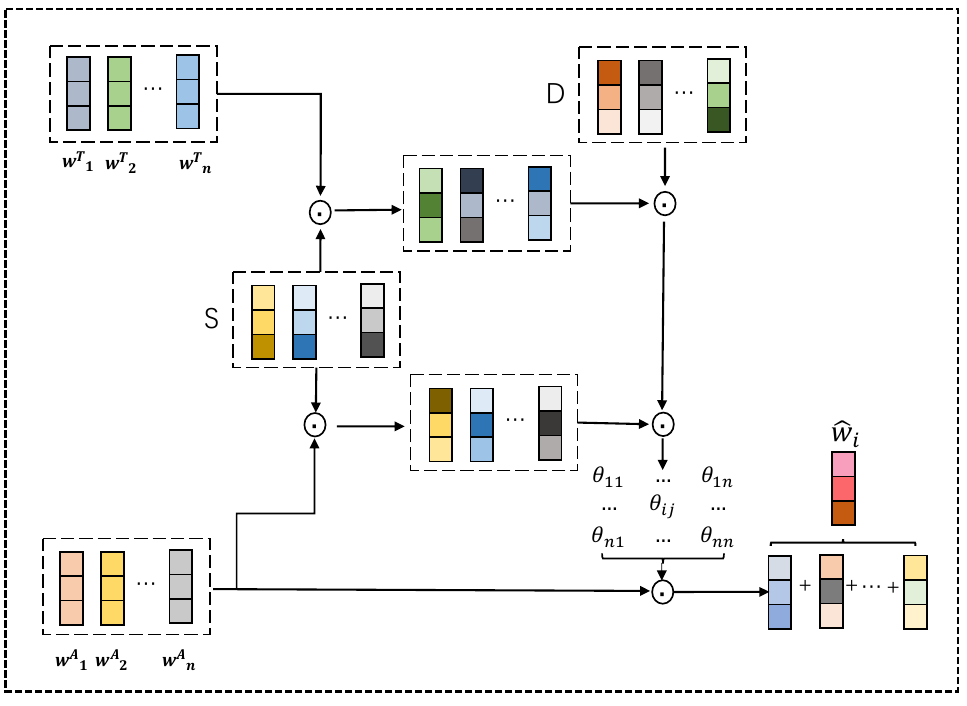}
    \caption{Architecture for Cross Attention Mechanism. $\odot$ is the element-wise multiplication.}
    \label{fig:att}
    % \vspace{-10pt}
\end{figure}

\myparagraph{Cross-Attention}
% \textbf{Cross-Attention.} \quad 
We conduct cross-attention to capture the relevance between the reference written documents and the corresponding ASR transcriptions. Given the representations of the reference text documents and the corresponding ASR transcriptions with $n$ tokens:~$\{\textbf{w}_1^T,...,\textbf{w}_n^{T}\}$ $\subset$ $\textbf{R}^d$ and $\{\textbf{w}_1^{A},...,\textbf{w}_n^{A}\}$ $\subset$ $\textbf{R}^d$, the attention function is computed as:
\begin{equation}
    H_{ij}=ReLU(S\textbf{w}_i^{T})\textit{D}ReLU(S\textbf{w}_j^{A})
\end{equation}
\begin{equation}
   \theta_{ij} \propto \textit{Exp}(H_{ij})
\end{equation}
\begin{equation}
   \hat{\textbf{w}}_i^{T}= \sum_j\theta_{ij}\textbf{w}_j^{A}
\end{equation}
where $S \in \textbf{R}^{d\times k}$, $D \in \textbf{R}^{k\times k}$ is a diagonal matrix, and $k$ denotes the attention hidden size. For brevity, we re-formulate the above word-level attention function as $\textbf{Attn}(\{\textbf{w}_i^T\}_{i=1}^{n},$
$\{\textbf{w}_i^A\}_{i=1}^{n},\{\textbf{w}_i^A\}_{i=1}^{n})$~(See in Figure \ref{fig:att}).

\myparagraph{RNN}
In order to obtain the better ASR-robust contextualized representations, we use the BiLSTM after the cross attention layer:  
% In this component, for the above cross-attention representations, we use one more layer of BiLSTM to obtain a high  level understanding.
\begin{equation}
   \tilde{\textbf{w}}_1^{T,L+1},...,\tilde{\textbf{w}}_n^{T,L+1}=BiLSTM(\hat{\textbf{w}}_1^{T},...,\hat{\textbf{w}}_n^{T})
\end{equation}
\begin{equation}
    \hat{\textbf{w}}_i^{T}=[\hat{\textbf{w}}_1^{T,1};...;\hat{\textbf{w}}_1^{T,L}]
\end{equation}
where $L$ denotes the number of RNN layers.

\myparagraph{Self-Attention}
Considering there exists a huge mismatch between ASR transcripts and the corresponding text documents, it is necessary to establish direct correlations between two transcriptions. In order to model the long-term dependency between two documents, we additionally introduce a self-attention layer to obtain self-attentive representations. The self-attentive vector can be computed as:
\begin{equation}
    \{\textbf{u}_i\}_{i=1}^{n}=\textbf{Attn}(\{\tilde{\textbf{w}}_i^T\}_{i=1}^{n},\{\tilde{\textbf{w}}_i^T\}_{i=1}^{n},\{\tilde{\textbf{w}}_i^T\}_{i=1}^{n}).
\end{equation}

\subsection{Knowledge Distillation}
\label{subsec:kd}
Inspired by the previous work~\cite{hinton2015distilling} that trains a~\textit{student} model to match the full softmax distribution of the~\textit{teacher} model, we adopt a~\textit{teacher-student} paradigm to distill ASR-robust knowledge into a single~\textit{student} CMRC model.

Let $z_{T}=\{\textbf{u}_i\}_{i=1}^{n}$ and $z_{S}=\{\textbf{w}_{i}^A\}_{i=1}^{n}$ denote the training input to the following~\textit{teacher} model and~\textit{student} model, respectively. $y=\{\textbf{A}_{i}\}_{i=1}^{n}$ denote corresponding sequence of gold labels. $\Psi_{S} = F_{S}(z_{S})$ and $\Psi_{T} = F_{T}(z_{T})$ are output score of potential~\textit{teacher} and~\textit{student} function, respectively. The final loss of the~\textit{student} is defined as:
% The condition probability is defined as: $p(y|x) = \sum_{i=1}^{n}$.
\begin{equation}
    L = \alpha L_{NLL}  (p_{\tau}(\Psi_{S}),p_{\tau}(\Psi_{T})) + (1 - \alpha) L_{KD}(\Psi_{T}, y)),
    % \sum_{x\in \mathcal{X}} (\alpha \tau^2 \mathcal{KL}(Phi_{\tau}(z_{S}), Phi_{\tau}(z_{T})) + (1 - \alpha) \mathcal{XE}(Phi_{T}, y) ),
\end{equation}
where $L_{NLL}$ and $L_{KD}$ are the negative log-likelihood loss and cross entropy loss, respectively. $p_{\tau}(\cdot)$ denotes the softmax function~\cite{guo2017calibration}. $\tau$ and $\alpha$ are hyperparameters.

\section{Experiments}
In this section, we first introduce several previous state-of-the-art language models. Then we investigate the performance of all evaluated baselines trained on Text-CoQA \cite{DBLP:journals/tacl/ReddyCM19} or Spoken-CoQA dataset. Finally, we demonstrate the effectiveness of the proposed CADNet.

\subsection{Baseline Models}
\myparagraph{FlowQA} \cite{huang2018flowqa} utilizes high-level granularity representations of questions and documents to enable the model to comprehend the topic of documents and integrate the underlying semantics of the dialogue history.

\myparagraph{SDNet}~\cite{zhu2018sdnet} uses inter-attention and self-attention to extract different levels of granularities to enable the model to understand the relevant information from passages and innovatively incorporate the BERT model.

\myparagraph{BERT-base} \cite{devlin2018bert} is a remarkable breakthrough in the natural language processing community, which achieves state-of-the-art performances in many downstream natural language processing tasks. It utilizes stacked transformers as encoders with residual structures.

\myparagraph{ALBERT-base}~\cite{lan2019albert}, a lightweight variant of BERT-based models, uses the parameter-sharing strategy in multiple parts of a model to compress the model size. It shares a similar structure with BERT, while maintaining comparable performances with smaller parameters.

\begin{table}[t]
    \caption{Comparison of model performance. We set the model on  text corpus as the~\textit{teacher} model, and the another on the ASR transcripts as the~\textit{student} model.}
    \centering
    \resizebox{0.98\columnwidth}{!}{%
    \begin{tabular}{l|cc|cc}
        % \toprule
%   \bottomrule[1pt]
    & \multicolumn{2}{|c|}{\textbf{T-CoQA dev}}&\multicolumn{2}{|c}{\textbf{S-CoQA test}}\\
    \hline\hline
    \textbf{Methods}&EM&F1 &EM&F1\\
    \hline
    FlowQA \cite{huang2018flowqa}&40.9&51.6&22.1&34.7\\
    FlowQA \cite{huang2018flowqa}+ sub-word unit \cite{li2018spoken}&41.9&53.2&23.3&36.4\\
    FlowQA \cite{huang2018flowqa}+ SLU \cite{serdyuk2018towards}&41.2&52.0&22.4&35.0\\
    FlowQA \cite{huang2018flowqa}+\textbf{CA} &41.6&52.6&23.2&35.8\\
    FlowQA \cite{huang2018flowqa}+\textbf{KD} &42.5&53.7&23.9&39.2\\
    FlowQA \cite{huang2018flowqa}+\textbf{CA+KD} &\textbf{43.2}&\textbf{54.7}&\textbf{25.0}&\textbf{40.3}\\
    \hline
    SDNet \cite{zhu2018sdnet}&40.1&52.5&41.5&53.1\\
    SDNet \cite{zhu2018sdnet}+ sub-word unit \cite{li2018spoken}&41.2&53.7&41.9&54.7\\
    SDNet \cite{zhu2018sdnet}+ SLU \cite{serdyuk2018towards}&40.2&52.9&41.7&53.2\\
    SDNet \cite{zhu2018sdnet}+\textbf{CA}&40.9&54.1&42.4&54.1\\
    SDNet \cite{zhu2018sdnet}+\textbf{KD}&41.7&55.6&43.6&56.7\\
    SDNet \cite{zhu2018sdnet}+\textbf{CA+KD}&\textbf{42.5}&\textbf{57.2}&\textbf{44.5}&\textbf{57.7}\\
    \hline
    BERT-base \cite{devlin2018bert}&42.3&55.8&40.6&54.1\\
    BERT-base \cite{devlin2018bert}+ sub-word unit \cite{li2018spoken}&43.2&56.8&41.6&55.4\\
    BERT-base+ SLU \cite{serdyuk2018towards}&42.5&56.1&41.0&54.6\\
    BERT-base \cite{devlin2018bert}+\textbf{CA}&43.3& 56.6& 41.6& 55.2\\
    BERT-base \cite{devlin2018bert}+\textbf{KD}&44.1& 58.8& 42.8& 57.7\\
    BERT-base \cite{devlin2018bert}+\textbf{CA+KD}&\textbf{45.1}& \textbf{59.9}& \textbf{43.8}& \textbf{58.9}\\
    \hline
    ALBERT-base \cite{lan2019albert}&42.7&56.0&41.4&55.2\\
    ALBERT-base \cite{lan2019albert}+ sub-word unit \cite{li2018spoken}&43.7&57.2&42.6&56.8\\
    ALBERT-base \cite{lan2019albert}+ SLU \cite{serdyuk2018towards}&42.8&56.3&41.7&55.7\\
    ALBERT-base \cite{lan2019albert}+\textbf{CA}&43.8&57.7&42.2&56.2\\
    ALBERT-base \cite{lan2019albert}+\textbf{KD}&44.8&59.6&43.9&58.7\\
    ALBERT-base \cite{lan2019albert}+\textbf{CA+KD}&\textbf{45.9}&\textbf{61.3}&\textbf{44.7}&\textbf{59.7}
    \\
    % \toprule
    % \bottomrule
%  \bottomrule[1pt]
    \end{tabular}
    }
    \label{tab:my_label_3}
\end{table}

\subsection{Experimental Settings}
We choose BERT-base and ALBERT-base in this study, consisting of 12 transformer encoders with hidden size 768. To guarantee the integrity of training, we follow the standard settings in four baselines. BERT and ALBERT utilize BPE as the tokenizer, but FlowQA and SDNet use SpaCy \cite{spacy2} for tokenization. Specifically, in the case that each token in spaCy corresponds to more than one BPE sub-tokens, we compute the embedding for each token by averaging the BERT embeddings of the relevant BPE sub-tokens. Note that we train all the evaluated baseline models over Text-CoQA in our local computing environment, which results in a little bit different from their results on the Text-CoQA leaderboard. In detail, we train FLowQA and SDNet for 30 epochs and fine-tune BERT and ALBERT for 5 epochs. We empirically set~$\alpha$ to 0.9 (defined in Section~\ref{subsec:kd}) in all experiments. We choose the maximum EM (Exact Match) and F1 score for evaluating the performance of SCQA models.

\subsection{Results}
In this study, we choose four state-of-the-art CMRC models (FlowQA~\cite{huang2018flowqa}, SDNet~\cite{zhu2018sdnet}, BERT-base~\cite{devlin2018bert}, ALBERT-base~\cite{lan2019albert}). We conduct two sets of experiments: 1). We train the baselines on Text-CoQA training set, and then evaluate the baselines on Text-CoQA dev set and Spoken-CoQA dev set, respectively; 2). We train the baselines on Spoken-CoQA training set and compare the baselines on Text-CoQA dev set and Spoken-CoQA test set, respectively.

In the first set of experiments, we report the results in Table~\ref{tab:my_label_2}. When looked into the table, we can find a large performance gap between the model trained on the text references (Text-CoQA training set) and another on ASR transcriptions (Spoken-CoQA training set), which indicates that recognition errors inevitably misled the CMRC models to make incorrect predictions. Therefore, it is desirable and meaningful to explore a more appropriate strategy to alleviate the adverse effects of ASR errors.

In the second set of experiments, Table~\ref{tab:my_label_3} reports the quantitative results. Our proposed~\textit{teacher-student} paradigm generally helps improve the baseline performance by injecting ASR-robust features into the baseline. With CA+KD, FlowQA achieves 54.7$\%$ (vs.51.6$\%$), and 39.9$\%$ (vs.34.7$\%$) on F1 score over the manual transcripts and ASR transcripts, respectively; SDNet outperforms the baseline model, achieving 56.5$\%$ (vs.52.5$\%$) and 57.7$\%$ (vs.53.1$\%$) on F1 score. As for two BERT-like models: BERT-base and ALBERT-base, our proposed KD strategy consistently leads to improvements over directly using baseline models, which achieves the F1 score of 58.8$\%$ (vs.55.8$\%$) and 57.7$\%$ (vs.54.1$\%$); 59.6$\%$ (vs.56.0$\%$) and 58.7$\%$ (vs.55.2$\%$). This demonstrates the effectiveness of our proposed KD strategy. We also evaluate the robustness of our CA mechanism. As shown in Table~\ref{tab:my_label_3}, our models trained with our CA mechanism further improve the results considerably. This suggests that utilizing CA can help the model comprehend the conversational context and extract relevant information from the passage, which is beneficial for final answer prediction.

\begin{figure}[h]
\centering    %居中
 
% \subfigure[] %第一张子图
{
	\begin{minipage}[h]{0.45\textwidth}
	\centering          %子图居中
	\includegraphics[width=0.9\columnwidth]{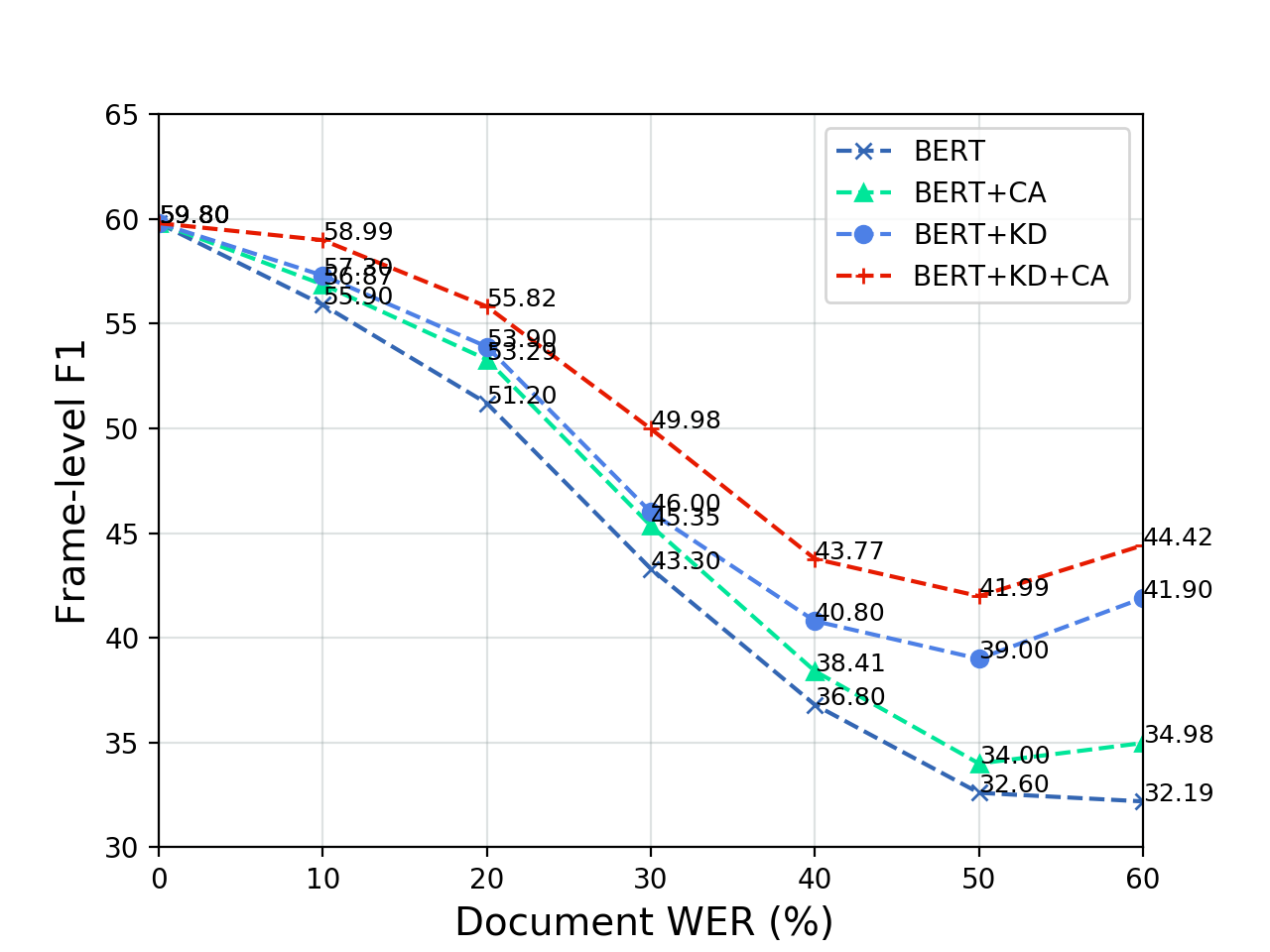}   %以pic.jpg的0.5倍大小输出
	\end{minipage}
}
	
% \subfigure[] %第二张子图
{
	\begin{minipage}[h]{0.45\textwidth}
	\centering      %子图居中
	\includegraphics[width=0.9\columnwidth]{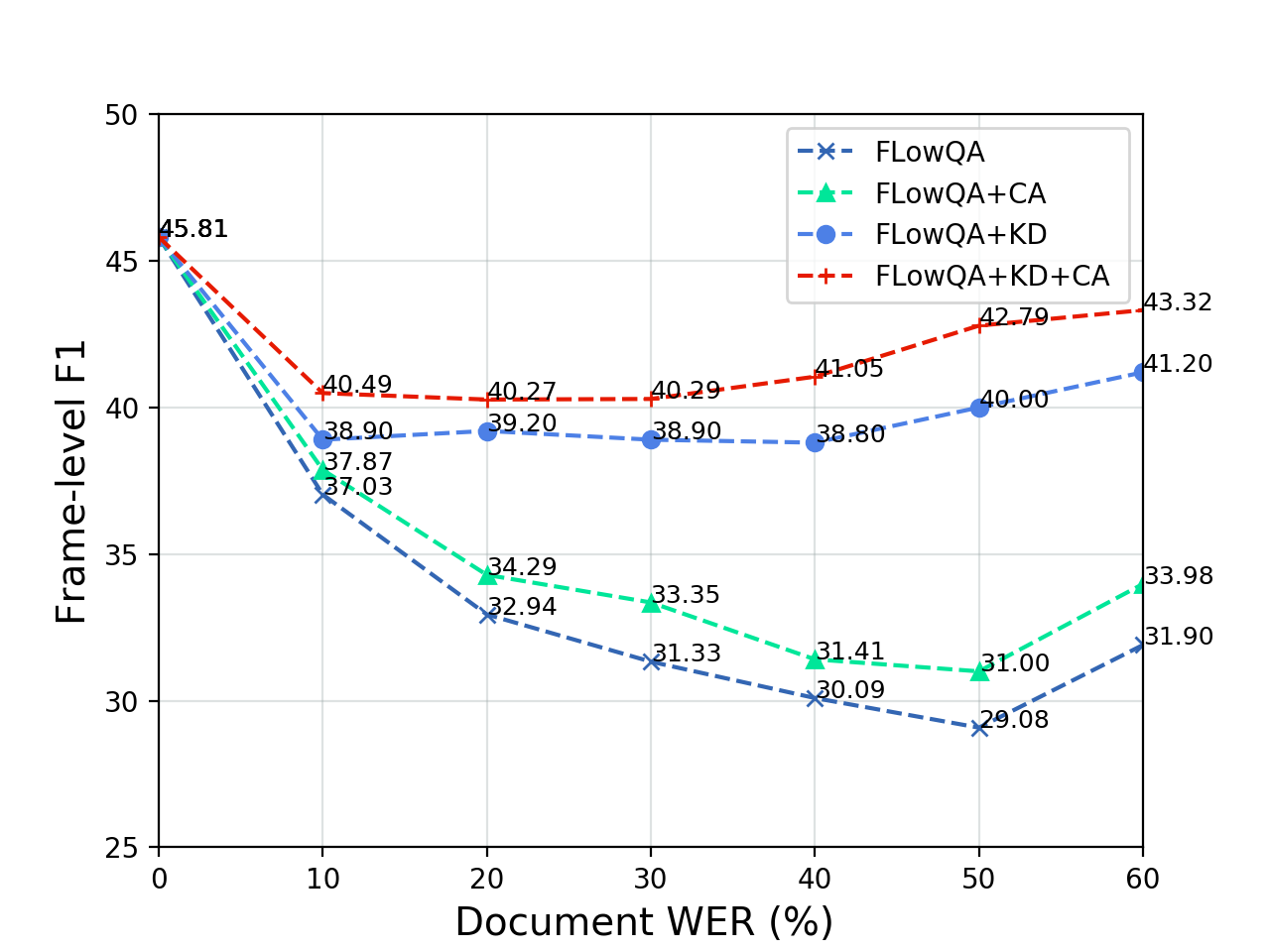}   %以pic.jpg的0.5倍大小输出
	\end{minipage}
}
 
\caption{Analysis of different WER on Spoken-CoQA.} %  %大图名称
\label{fig:wer}  %图片引用标记
\end{figure}

To further illustrate the generalization of the proposed approach, we conduct experiments to investigate model performance on speech documents with different Word Error Rates (WERs). We first split Spoken-CoQA into small sub-sets with different WERs. We then utilize Frame-level F1 score \cite{chuang2019speechbert} as an additional evaluation metric to validate our proposed method on Spoken-CoQA. In Figure \ref{fig:wer}, we can learn that all the methods (FlowQA and BERT) have a significant drop in performance at higher WER. We observe that sequentially using the knowledge distillation strategy and contextualized attention mechanism are capable of yielding considerable performance boosts on all the evaluated models. This suggests that adopting these two strategies on SCQA tasks is able to improve network performance at higher WER.

\section{Conclusion}
In this paper, we propose a contextual attention-based knowledge distillation network, CADNet, to tackle the spoken conversational question answering task. We propose to leverage cross-attention and self-attention on both manual and ASR transcriptions, and employ a~\textit{teacher-student} framework to distill ASR-robust contextualized knowledge into the~\textit{student} model. Experimental results show that our method outperforms all the evaluated baselines and further mitigates the negative effects of ASR errors. Our future work is to explore more effective speech and text fusing mechanisms to further improve performance.

\section{Acknowledgements}
This paper is partially supported by Shenzhen municipal research funding project and Technology Fundamental Research Programs (No: GXWD20201231165807007-20200814115301001 \& JSGG20191129105421211).

\bibliographystyle{IEEEtran}

\bibliography{mybib}

\end{document}